\documentclass[runningheads]{llncs}
\usepackage[T1]{fontenc}
\usepackage{graphicx}
\usepackage{amsmath,amssymb}
\usepackage{booktabs}
\usepackage{multirow}
\usepackage{url}
\usepackage{cite}
\usepackage{hyperref}
\usepackage{algorithm}
\usepackage{algpseudocode}
\usepackage{cleveref}
\usepackage[protrusion=false]{microtype}
\begin{document}
\title{Hindsight Preference Replay Improves Preference-Conditioned Multi-Objective Reinforcement Learning}
\titlerunning{Hindsight Preference Replay for Multi-Objective RL}
%
\author{Jonaid Shianifar\inst{1}\orcidID{0000-0003-0477-0056} \and
Michael Schukat\inst{1}\orcidID{0000-0002-6908-6100} \and
Karl Mason\inst{1}\orcidID{0000-0002-8966-9100}}

\authorrunning{J. Shianifar et al.}
\institute{School of Computer Science, University of Galway, H91 FYH2, Galway, Ireland 
\email{\{J.Shianifar1,Michael.Schukat,Karl.Mason\}@universityofgalway.ie}}

\maketitle              
\microtypesetup{protrusion=false}
\begingroup\emergencystretch=2em
\begin{abstract}

Multi-objective reinforcement learning (MORL) enables agents to optimize vector-valued rewards while respecting user preferences. CAPQL, a preference-conditioned actor-critic method, achieves this by conditioning on weight vectors $\mathbf{w}$ and restricts data usage to the specific preferences under which it was collected, leaving off-policy data from other preferences unused. We introduce Hindsight Preference Replay (HPR), a simple and general replay augmentation strategy that retroactively relabels stored transitions with alternative preferences. This densifies supervision across the preference simplex without altering the CAPQL architecture or loss functions. Evaluated on six MO-Gymnasium locomotion tasks at a fixed 300{,}000-step budget using expected utility (EUM), hypervolume (HV), and sparsity, HPR\mbox{-}CAPQL improves HV in five of six environments and EUM in four of six. On \texttt{mo-humanoid-v5}, for instance, EUM rises from $323\!\pm\!125$ to $1613\!\pm\!464$ and HV from $0.52$M to $9.63$M, with strong statistical support. \texttt{mo-halfcheetah-v5} remains a challenging exception where CAPQL attains higher HV at comparable EUM. We report final summaries and Pareto-front visualizations across all tasks.\footnote{This paper has been accepted for presentation at the 33rd Irish Conference on Artificial Intelligence and Cognitive Science (AICS 2025).}


\end{abstract}

\section{Introduction}
Many real-world control tasks, such as robotic locomotion or autonomous driving, involve optimizing multiple conflicting objectives, for example, energy efficiency versus speed, or safety versus exploration. Multi-objective reinforcement learning (MORL) formalizes this by optimizing vector-valued rewards $\mathbf{r}_t \in \mathbb{R}^m$ \cite{zhang2023multi, van2014multi}. However, selecting a single solution requires expressing user preferences over these objectives \cite{hayes2021practical}.

One approach to MORL is scalarization, where a weight vector $\mathbf{w} \in \Delta^m$ is used to linearly combine the objectives: $u_\mathbf{w}(\mathbf{r}) = \mathbf{w}^\top \mathbf{r}$. Preference-conditioned policies exploit this idea by taking $\mathbf{w}$ as input, enabling a single policy to serve a range of trade-offs depending on user preferences. This setup supports efficient deployment in dynamic or interactive systems where preferences may change over time \cite{lu2023multi, roijers2013survey}.

CAPQL~\cite{lu2023multi} is a recent preference-conditioned actor-critic method that extends Soft Actor-Critic (SAC)~\cite{haarnoja2018soft} to the MORL setting. CAPQL augments the critic loss with a concavity-based regularizer to improve coverage and training stability. However, like most off-policy algorithms, it suffers from a key inefficiency: data collected under one preference $\mathbf{w}$ is only used to train for that same $\mathbf{w}$ \cite{lu2023multi}.

This results in low sample efficiency, as transitions informative for neighboring preferences are discarded. This inefficiency is especially costly in high-dimensional control tasks, where sample collection is expensive. Inspired by Hindsight Experience Replay (HER) that relabels goals to overcome sparse rewards~\cite{andrychowicz2017hindsight, wan2018advances}, we propose \textit{Hindsight Preference Replay} (HPR), which retroactively relabels preference vectors, densifying supervision without architectural or loss changes~\cite{fan2025preference}. Empirically, HPR-CAPQL improves hypervolume in five of six benchmarks, with especially large gains on \texttt{mo-humanoid-v5}, while \texttt{mo-halfcheetah-v5} remains a challenging case where CAPQL covers the front more broadly.

Our contributions are as follows:
\begin{itemize}
  \item We present HPR, a general-purpose hindsight relabeling mechanism for preference-conditioned MORL.
  \item We instantiate HPR within CAPQL, creating HPR-CAPQL, with minimal implementation overhead.
  \item We evaluate HPR-CAPQL on six continuous control tasks from the MO-Gymnasium benchmark suite, showing performance gains in several environments.
\end{itemize}

\section{Background and Literature Review}
Modern control problems, such as robotic manipulation and locomotion, inherently require optimizing multiple conflicting objectives, for example, achieving both high efficiency and safety. The field of multi-objective reinforcement learning (MORL) formalizes this by extending classical reinforcement learning to vector-valued rewards $\mathbf{r}_t \in \mathbb{R}^m$, enabling agents to discover policies that provide favorable trade-offs across the Pareto front \cite{roijers2013survey, van2014multi, zitzler2007hypervolume, mogymnasium}.

A common approach is \emph{scalarization}, where a preference or weight vector $\mathbf{w}\in\Delta^{m}$ combines objectives into a single utility $u_{\mathbf{w}}(\mathbf{r})=\mathbf{w}^{\top}\mathbf{r}$, thereby parameterizing the trade-off space~\cite{roijers2013survey}. Recent progress on adaptive scalarization reports improved stability and Pareto coverage in MORL, particularly in continuous-control settings~\cite{shianifar2025adaptive}. In parallel, preference-conditioned policies condition both policy and value networks on $\mathbf{w}$ to adapt to user preferences and efficiently cover the Pareto front~\cite{lu2023multi,mogymnasium}. From a robotics perspective, deep RL for adaptive robotic arm control underscores the need for sample-efficient replay and safe, budgeted interaction~\cite{shianifar2024optimizing}. Our approach is complementary to these lines: HPR augments replay to reuse trajectories across preferences without modifying the underlying losses or architectures.

Among these methods, \emph{CAPQL} (Concavity-Aware Preference-based Q-Learning) has emerged as a strong baseline in preference-conditioned MORL. CAPQL extends the Soft Actor-Critic (SAC) framework by conditioning both policy and critics on $\mathbf{w}$, enabling a single policy to generalize across multiple trade-offs. A key innovation in CAPQL is a concavity-based regularizer in the critic loss, which encourages stable training and broad coverage of the Pareto front by penalizing deviations from concave utility approximations. This design improves sample efficiency compared to earlier preference-conditioned approaches; however, CAPQL still updates policies only for the specific preferences under which data were collected, leaving off-policy experience from other preferences unused \cite{lu2023multi}.

A milestone in single-objective, goal-conditioned RL was the introduction of \emph{Hindsight Experience Replay} (HER), which relabels failed transitions with alternative goals to improve learning under sparse rewards \cite{andrychowicz2017hindsight}. HER established that retroactive relabeling can transform unsuccessful experiences into effective learning signals. Inspired by this success, researchers have explored extending hindsight mechanisms to other conditioning variables such as preferences.

The recently proposed \emph{Hindsight Preference Replay} (HPR) adapts HER’s philosophy to the multi-objective setting by retroactively relabeling the preference vector $\mathbf{w}$ in stored transitions. This enables richer and denser supervision across the preference simplex without any architectural or loss modification, offering large improvements in sample efficiency and Pareto coverage when combined with preference-conditioned methods like CAPQL \cite{lu2023multi}.

Earlier work on multi-objective HER primarily focused on relabeling goals and generating curriculum for complex robotic tasks using evolutionary algorithms \cite{andrychowicz2017hindsight, mogymnasium}, but did not address the inefficiency of preference-conditioned experience reuse. Recent strategies in multi-objective learning, such as Ensemble Multi-Objective RL (EMORL) \cite{mogymnasium}, curriculum generation via evolutionary multi-objective optimization \cite{andrychowicz2017hindsight,zitzler2007hypervolume, auger2009theory}, and meta-learning for adaptation, have pushed the boundaries of scalable multi-objective RL; however, they do not apply retroactive preference relabeling.

The proposed HPR therefore offers a novel and practical mechanism to improve sample efficiency and generalization in preference-conditioned MORL, building on the proven utility of hindsight-based replay and extending it in previously unexplored directions.

\section{Methodology: Hindsight Preference Replay (HPR)}
In preference-conditioned MORL, transitions collected under a specific preference vector $\mathbf{w}$ are typically used exclusively to update the policy for that same $\mathbf{w}$. However, these transitions may also be informative for other preferences, particularly those nearby in the simplex. This leads to inefficiencies in data usage, limiting gradient diversity and generalization.

To address this, we propose \textit{Hindsight Preference Replay} (HPR), a replay augmentation strategy that retroactively relabels transitions with alternative preference vectors. The key idea is analogous to HER~\cite{andrychowicz2017hindsight}, but adapted to preference relabeling rather than goal relabeling.

The pseudocode in Algorithm~\ref{alg:hpr-capql} summarizes the full HPR-CAPQL procedure integrated with the CAPQL-style actor-critic learning. It highlights how transitions are collected and relabeled with multiple alternative preferences, stored in the replay buffer, and how training updates interleave with environment interaction and evaluation. This overview guides the detailed design of the relabeling strategies and integration steps discussed next.

\begin{algorithm}[H]
\caption{HPR-CAPQL: Hindsight Preference Replay for Preference-Conditioned MORL}
\label{alg:hpr-capql}
\begin{algorithmic}[1]
\Require Environment $Env$; preference-conditioned agent (policy/critics conditioned on $w$); replay buffer $\mathcal{B}$;
per-step relabels $K$; Dirichlet concentration $\kappa$; relabeled minibatch fraction $\rho$; (optional) acceptance test.
\State Initialize agent and targets; $\mathcal{B}\leftarrow\emptyset$
\For{episodes until budget exhausted}
  \State Sample behavior preference $w \sim p(w)$ on the simplex; reset $Env$; set episode return $G\!\leftarrow\!{\bf 0}$
  \While{not done}
     \State Act $a\!\sim\!\pi(\cdot\mid s,w)$, step env $\rightarrow (r,s',d)$; push $(s,a,r,s',d,w)$ to $\mathcal{B}$
     \State $G \leftarrow G + \gamma^t r$ \Comment{\emph{discounted vector return for relabeling/eval}}
     \For{$k=1..K$} \Comment{\emph{Neighborhood relabeling}}
        \State $\tilde w \sim \mathrm{Dir}(\kappa w)$; \textbf{if} accepted \textbf{then} push $(s,a,r,s',d,\tilde w)$ to $B$
     \EndFor
     \State $s \leftarrow s'$
     \State \textbf{Training step(s):} sample minibatches from $B$ with relabeled fraction $\rho$; update CAPQL with scalarized rewards $r_w = w^\top r$
  \EndWhile
  \State \Comment{\emph{Return-aligned relabeling (episode end)}}
  \State $\hat w \leftarrow \Pi_{\Delta}\!\big(\mathrm{softplus}(G)\big)$; for each episode transition push $(s,a,r,s',d,\hat w)$ if accepted
\EndFor
\State \textbf{Evaluation:} on a fixed weight grid, log EUM, HV, sparsity, and per-objective means
\end{algorithmic}
\end{algorithm}

Given a stored transition tuple $(s, a, \mathbf{r}, s', \mathbf{w})$, HPR replaces the original preference vector $\mathbf{w}$ with a new preference vector $\tilde{\mathbf{w}}$ sampled by return-aligned or neighborhood strategies~\cite{fan2025preference, yangpreference}. The reward vector $\mathbf{r}$ is left unchanged, enabling recomputation of scalarized returns $u_{\tilde{\mathbf{w}}}(\mathbf{r}) = \tilde{\mathbf{w}}^\top \mathbf{r}$ and allowing a single transition to serve multiple scalarization preferences. This approach generalizes the HER principle~\cite{andrychowicz2017hindsight, wan2018advances} into the preference space, enabling richer supervision without modifying CAPQL’s architecture~\cite{lu2023multi}.

\subsection{Relabeling Strategies}
We implement two relabeling strategies in HPR:

\paragraph{\textbf{Return-Aligned Relabeling.}}
For each trajectory or transition, we compute a discounted cumulative return vector $\mathbf{G}$. We then define the relabeled preference $\tilde{\mathbf{w}}$ as:
\begin{equation}
    \tilde{\mathbf{w}} \propto \text{softplus}(\mathbf{G})
\end{equation}

followed by projection onto the simplex. Optionally, a convex combination with the original $\mathbf{w}$ can be applied to control deviation. This heuristic aligns the relabeling with the outcomes the trajectory has already achieved.

\paragraph{\textbf{Neighborhood Sampling.}}
We draw $K$ new preference vectors $\tilde{\mathbf{w}} \sim \text{Dir}(\kappa \mathbf{w})$ from a Dirichlet distribution centered around the original $\mathbf{w}$, where $\kappa$ is a concentration parameter. Larger $\kappa$ values result in narrower neighborhoods, encouraging local generalization.

\subsection{Integration with CAPQL}
We integrate HPR into CAPQL by modifying the replay buffer insertion logic and minibatch sampling:

\begin{itemize}
  \item During data collection, each transition is stored as $(s, a, \mathbf{r}, s', \mathbf{w})$.
  \item On insertion (or periodically), up to $K$ relabeled versions are added using one of the above strategies, with preferences $\tilde{\mathbf{w}}$.
  \item During training, minibatches are sampled uniformly, mixing original and relabeled transitions.
\end{itemize}

Importantly, all CAPQL loss functions and network structures remain unchanged. HPR only alters the diversity of training data in the preference space.

\subsection{Practical Settings and Filters}
We explore relabel counts $K \in \{0, 1, 2, 4\}$, Dirichlet concentration parameters $\kappa \in \{10, 20, 50\}$, and relabel batch ratios $\rho \in [0.3, 0.7]$. 

In settings with strongly orthogonal objectives, naive relabeling can harm learning. In such cases, we apply acceptance filters based on alignment between $\tilde{w}$ and $G$, e.g., $\cos(\tilde{w},G) \ge \tau$ or $\tilde{w}^\top G \ge w^\top G - \varepsilon$, with $\varepsilon \in \{0.0, 0.1\}$ and $\tau \in \{0.7, 0.8\}$. This mitigates regressions in environments where broad relabels destabilize HV (e.g., \texttt{mo-halfcheetah-v5}), while preserving the gains observed elsewhere.

\section{Experimental Setup}

\subsection{Environments}

We evaluate HPR-CAPQL across six continuous control tasks from the MO-Gymnasium benchmark suite~\cite{mogymnasium}, an extension of Gymnasium~\cite{gymnasium} designed for multi-objective reinforcement learning. All environments feature continuous observations and continuous action spaces, returning 2D reward vectors that correspond to conflicting objectives.

The environments are described as follows:

\begin{itemize}
  \item \textbf{mo-hopper-2obj-v5}: A 2D hopping robot optimizing for forward progress and jumping height.
  \item \textbf{mo-walker2d-v5}: A bipedal walker balancing locomotion efficiency and energy cost.
  \item \textbf{mo-halfcheetah-v5}: A planar cheetah optimizing for speed versus energy efficiency.
  \item \textbf{mo-humanoid-v5}: A high-dimensional humanoid robot focusing on fast movement and minimal effort.
  \item \textbf{mo-swimmer-v5}: A simple two-link swimmer navigating through fluid.
  \item \textbf{mo-ant-2obj-v5}: A quadruped ant robot optimizing its movement in both X and Y directions.
\end{itemize}

Key characteristics of these environments, including observation dimensions, action dimensions, and reward components, are summarized in Table~\ref{tab:env_specs}.

\begin{table}[ht]
\centering
\caption{Observation and Action Dimensions and Reward Components of MO-Gymnasium Environments}
\label{tab:env_specs}
\begin{tabular}{lccc}
\toprule
Environment & Observation Dim. & Action Dim. & Reward Components \\
\midrule
mo-hopper-2obj-v5 & 11 & 3 & Forward velocity, Jump height \\
mo-walker2d-v5 & 17 & 6 & Forward velocity, Control cost \\
mo-halfcheetah-v5 & 17 & 6 & Forward velocity, Control cost \\
mo-humanoid-v5 & 348 & 17 & Forward velocity, Control cost \\
mo-swimmer-v5 & 8 & 2 & Forward velocity, Control cost \\
mo-ant-2obj-v5 & 105 & 8 & X-velocity, Y-velocity \\
\bottomrule
\end{tabular}
\end{table}

All environments utilize MuJoCo physics and provide dense, continuous rewards.

\subsection{Training Protocol}
We compare standard CAPQL against HPR-CAPQL using the same codebase for both methods. Each algorithm is trained for $300{,}000$ environment steps per task, with evaluation every $10{,}000$ steps on a fixed preference grid over the 2D simplex. We report the mean and standard deviation over five seeds. Metrics include Expected Utility (EUM), Hypervolume (HV), and the sparsity of the nondominated archive; we also plot the final Pareto fronts. HV is computed with a fixed per-environment reference point $\,\mathbf{r}_{\mathrm{ref}}=[-100,-100]\,$ shared by all methods and seeds, and we report HV in millions $(\times 10^6)$ for readability.
\subsection{Hardware Configuration}
All experiments were conducted on a high-performance computing server equipped with an AMD Ryzen Threadripper 7960X 24-core processor, two NVIDIA RTX A6000 GPUs, and 128 GB of system RAM. The system ran Ubuntu 22.04.5 LTS. The software stack included Python 3.9.23, PyTorch 2.8.0, Gymnasium 1.1.1, and MO-Gymnasium 1.3.1. This configuration provided ample computational power for efficient training and evaluation of preference-conditioned policies across all benchmark tasks.

\section{Results and Discussion}
We evaluate CAPQL and HPR-CAPQL for $300{,}000$ environment steps on six MO-Gymnasium tasks. We report Expected Utility (EUM; $\uparrow$), hypervolume (HV; $\uparrow$; Table values in millions; For each environment, we use a fixed $r_{ref}=[-100, -100]$), and sparsity (archive spacing; $\downarrow$). Final summaries with Welch tests for HV appear in Table~\ref{tab:final-results}. \Cref{fig:eum-curves,fig:hv-curves,fig:Sparsity-curves} show EUM/HV/sparsity learning curves, and Figure~\ref{fig:pareto} visualizes final Pareto fronts from the union of non-dominated points across seeds.

\subsection{Final Performance Summary}
Across tasks, hindsight preference relabeling primarily increases Pareto coverage. HPR-CAPQL attains higher HV in most environments and improves EUM in a majority as well; the aggregate row of Table~\ref{tab:final-results} reflects this pattern. Several HV differences are statistically reliable, notably on \texttt{mo-humanoid-v5}, \texttt{mo-walker2d-v5}, and \texttt{mo-ant-2obj-v5}. Two tasks deviate from this trend. \texttt{mo-halfcheetah-v5} favors CAPQL on HV while EUM remains comparable, indicating that additional relabels do not widen the front in that domain. \texttt{mo-swimmer-v5} is effectively a draw: both algorithms saturate quickly and the small HV differences are not significant.

\begin{table}[ht]
\centering
\caption{Final performance after $300{,}000$ steps (mean $\pm$ std over seeds). HV shown in millions ($\times 10^6$) using the fixed $\mathbf{r}_{\mathrm{ref}}$. Significance markers: $^{*}$ for $p<0.05$, $^{**}$ for $p<0.01$, and $^{***}$ for $p<0.001$.}
\label{tab:final-results}
\begin{tabular}{l l l l l l}
\hline
Environment& Algorithm & EUM $\uparrow$& Sparsity $\downarrow$ & HV $\uparrow$ & HV $p$-value\\ \hline
\multirow{2}{*}{mo-hopper-2obj} & CAPQL&2370.5$\pm$634.7 & \textbf{84.6$\pm$76.8 }& 4.7534$\pm$2.39&\multirow{2}{*}{0.0749} \\
& HPR-CAPQL &\textbf{2942.7$\pm$143.8}&295.9$\pm$81.5&\textbf{7.2026$\pm$1.03}&\\ \hline
\multirow{2}{*}{mo-walker2d} & CAPQL&1056.4$\pm$273.3&2$294.8\pm$230.0&1.3276$\pm$0.52&\multirow{2}{*}{0.0079 **} \\
& HPR-CAPQL &\textbf{1320.7$\pm$207.6}&\textbf{513.5$\pm$361.5}&\textbf{2.9855$\pm$0.91}&\\ \hline
\multirow{2}{*}{mo-halfcheetah} & CAPQL&\textbf{1616.7$\pm$202.2}&\textbf{416.3$\pm$116.6}&\textbf{8.7208$\pm$1.48}&\multirow{2}{*}{0.0350 *} \\
& HPR-CAPQL &1603.0$\pm$319.7&535.9$\pm$183.1&5.9927$\pm$1.97&\\ \hline
\multirow{2}{*}{mo-humanoid} & CAPQL&323.1$\pm$125.4&\textbf{120.0$\pm$44.3}&0.5173$\pm$0.27&\multirow{2}{*}{0.0009 ***} \\
& HPR-CAPQL &\textbf{1613.3$\pm$464.3}&486.5$\pm$88.1&\textbf{9.6332$\pm$2.94}&\\ \hline
\multirow{2}{*}{mo-ant-2obj} & CAPQL&1823.3$\pm$191.2&\textbf{265.9$\pm$58.2}&3.2311$\pm$0.31&\multirow{2}{*}{0.0060 **} \\
& HPR-CAPQL &\textbf{2060.7$\pm$157.1}&653.6$\pm$356.2&\textbf{4.5428$\pm$0.68}&\\ \hline
\multirow{2}{*}{mo-swimmer} & CAPQL&\textbf{22.2$\pm$1.7}&10.3$\pm$7.1&0.0054$\pm$0.0030&\multirow{2}{*}{0.2926} \\
& HPR-CAPQL &22.1$\pm$1.9&\textbf{5.9$\pm$1.3}&\textbf{0.0104$\pm$0.0090}&\\ \hline
\hline
\multicolumn{2}{r}{\textbf{HPR-CAPQL better (of 6)}} & \textbf{4} & \textbf{2} & \textbf{5} & \textbf{—} \\
\hline
\end{tabular}
\end{table}

\subsection{Learning Behavior Over Time}
\Cref{fig:eum-curves,fig:hv-curves,fig:Sparsity-curves} plot EUM, HV, and sparsity over the $300{,}000$-step horizon and show how the final differences in Table~\ref{tab:final-results} emerge during training.
 On \texttt{mo-hopper-2obj-v5} and \texttt{mo-ant-2obj-v5}, HPR-CAPQL accelerates early, establishing EUM and HV margins that persist through the end of training. On \texttt{mo-humanoid-v5}, gains arrive later but become pronounced: once the policy stabilizes, relabeled preferences unlock new parts of the frontier and HV separates sharply. \texttt{mo-walker2d-v5} shows the same qualitative pattern with narrower uncertainty. In contrast, \texttt{mo-halfcheetah-v5} exhibits a plateau for HPR-CAPQL while CAPQL continues to expand HV, and \texttt{mo-swimmer-v5} reaches a quick equilibrium for both methods. Variance bands widen in the tasks where coverage expands the most, reflecting a broader exploration of the preference simplex rather than instability of the learner.
 
\begin{figure*}[hbt!]
  \centering
  \includegraphics[width=0.32\textwidth]{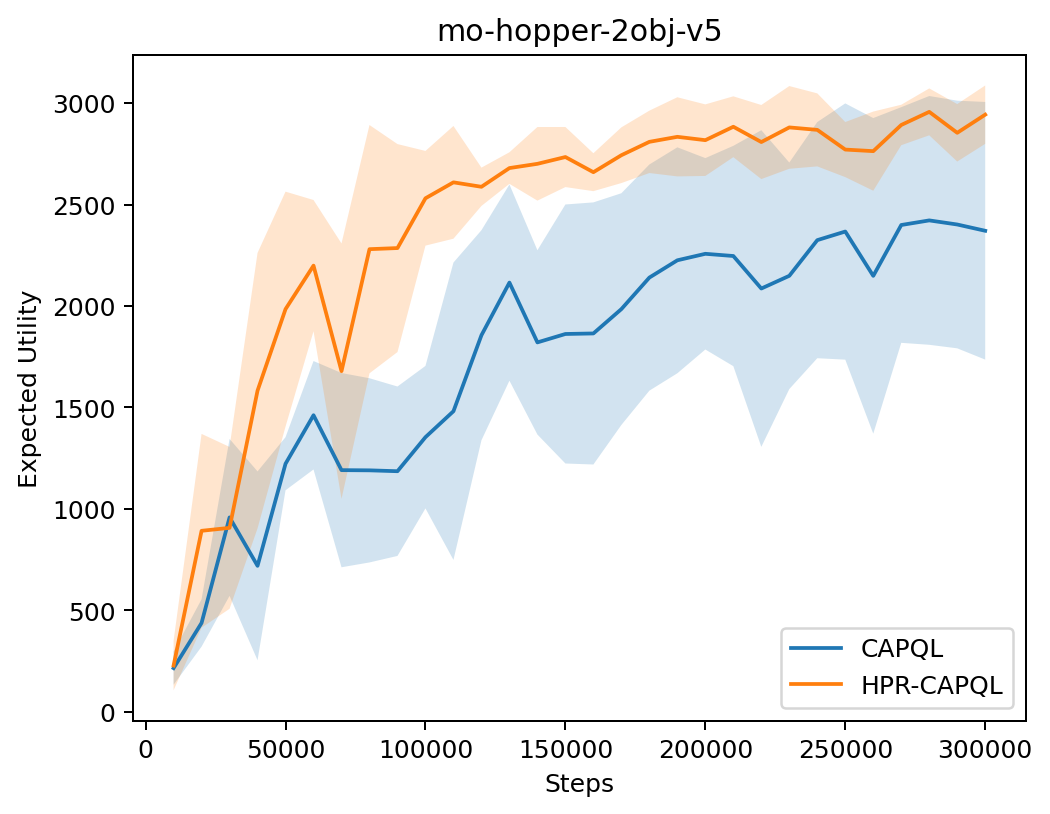}\hfill
  \includegraphics[width=0.32\textwidth]{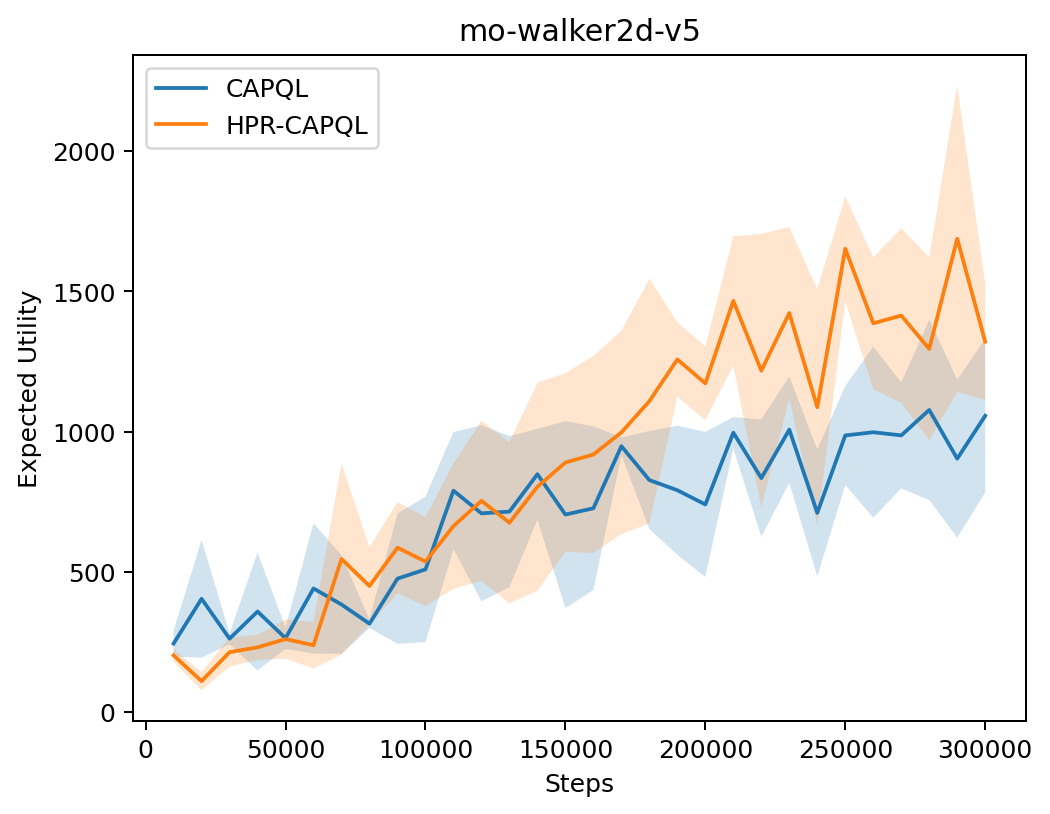}\hfill
  \includegraphics[width=0.32\textwidth]{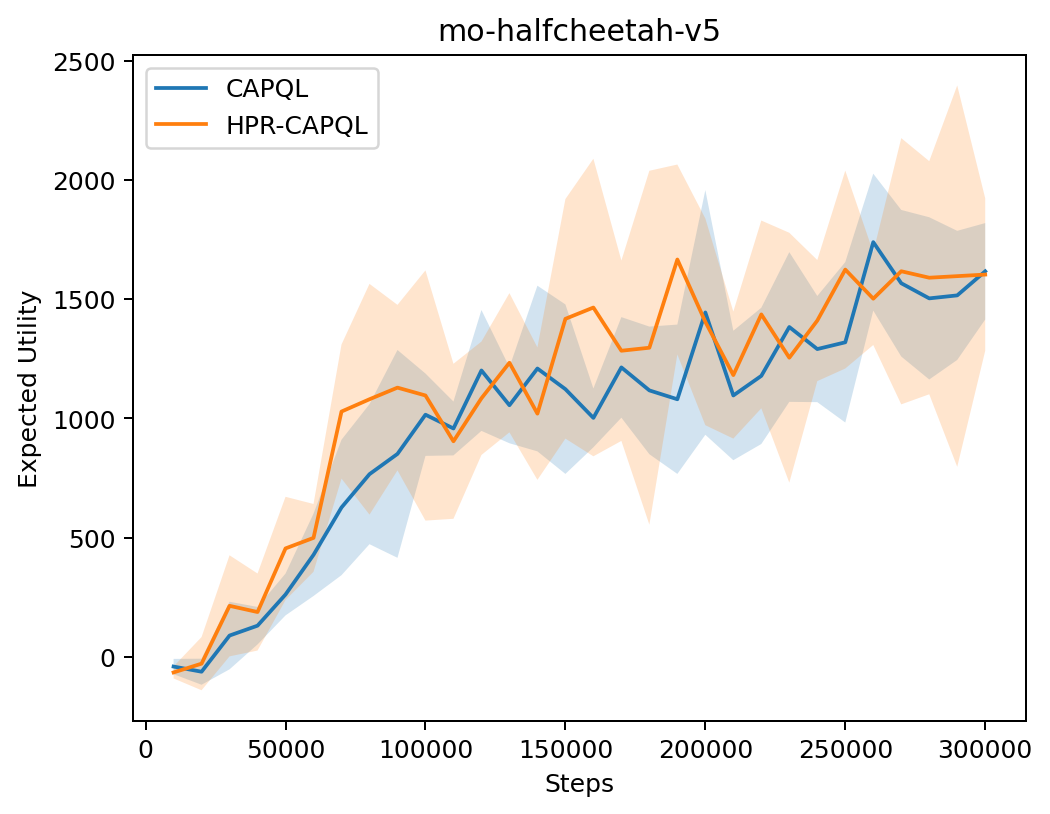}\\[0.6em]
  \includegraphics[width=0.32\textwidth]{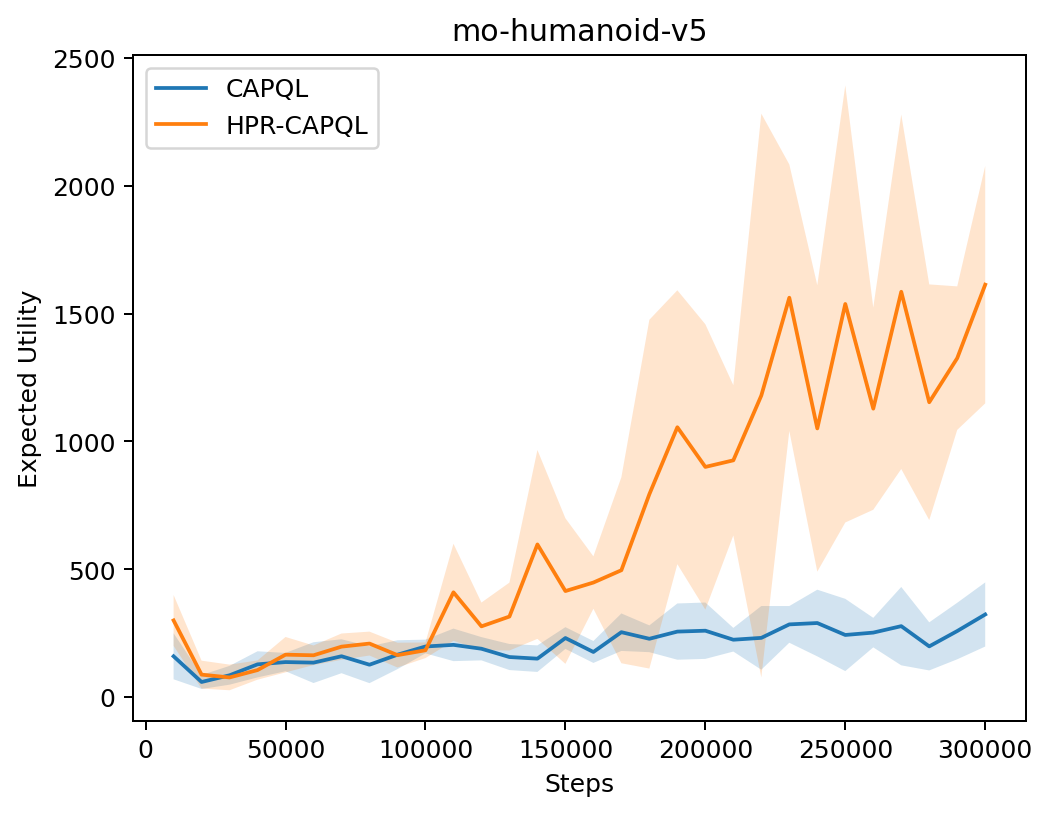}\hfill
  \includegraphics[width=0.32\textwidth]{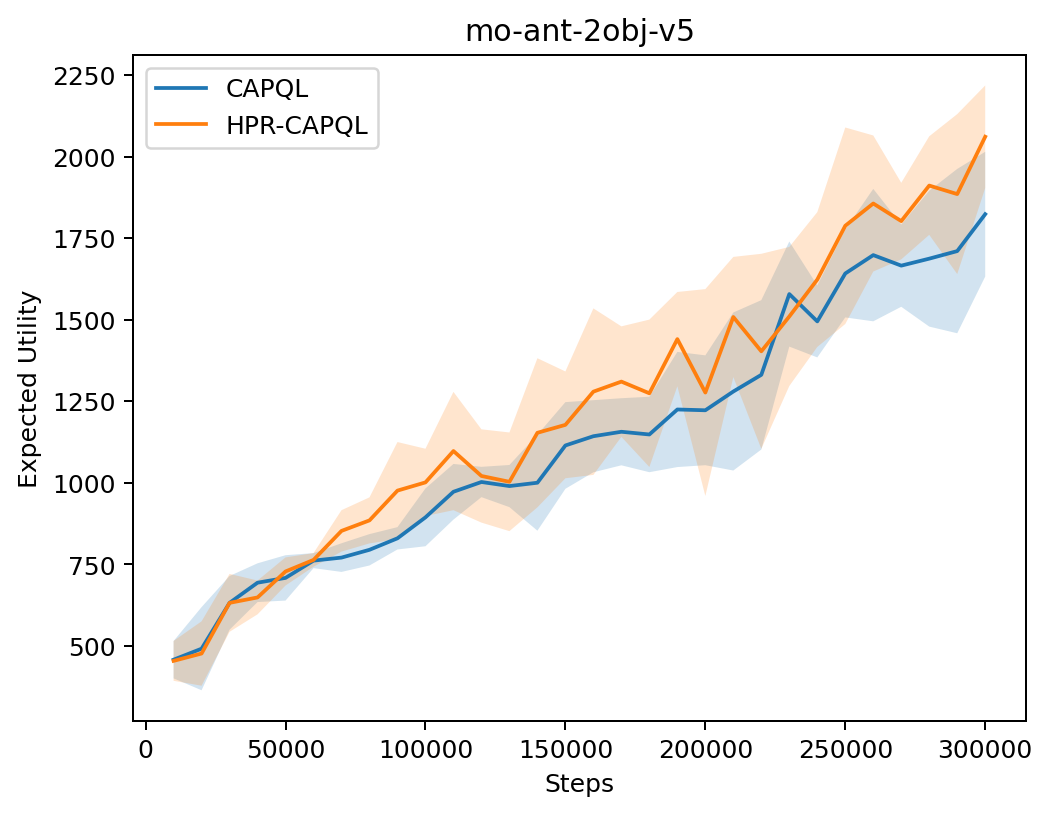}\hfill
  \includegraphics[width=0.32\textwidth]{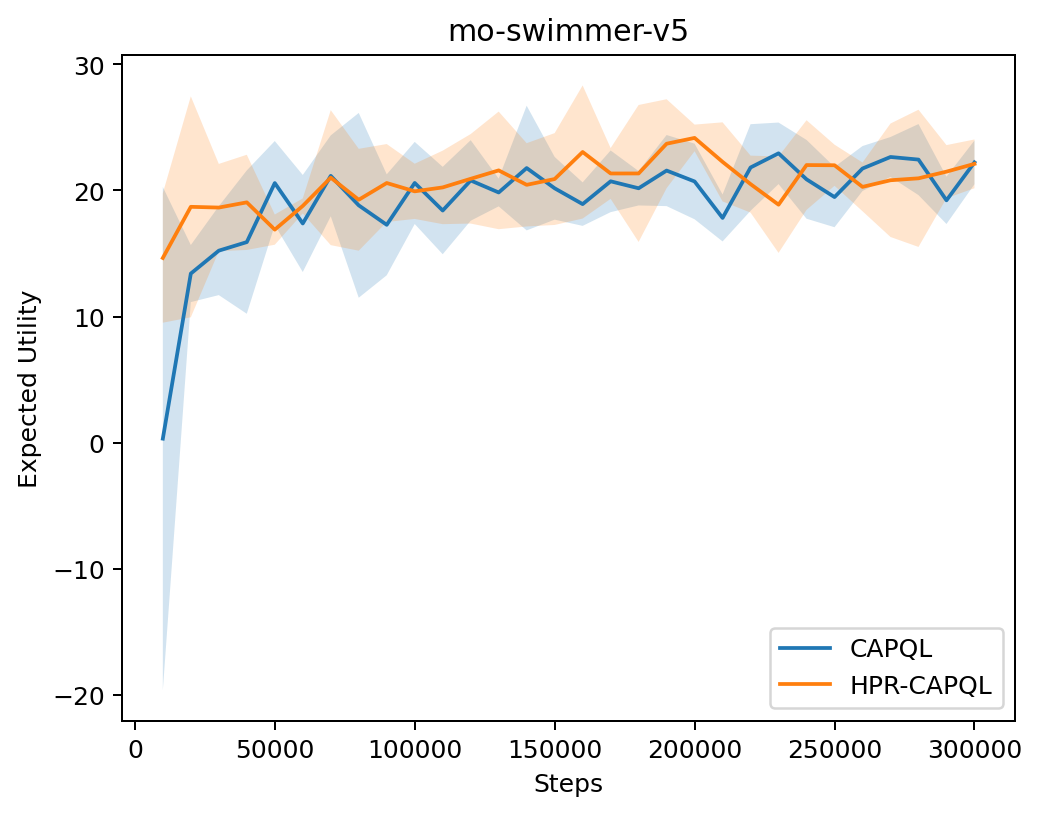}
  \caption{EUM learning curves up to $300{,}000$ steps (mean $\pm$ std over five seeds).}
  \label{fig:eum-curves}
\end{figure*}
\begin{figure*}[hbt!]
  \centering
  \includegraphics[width=0.32\textwidth]{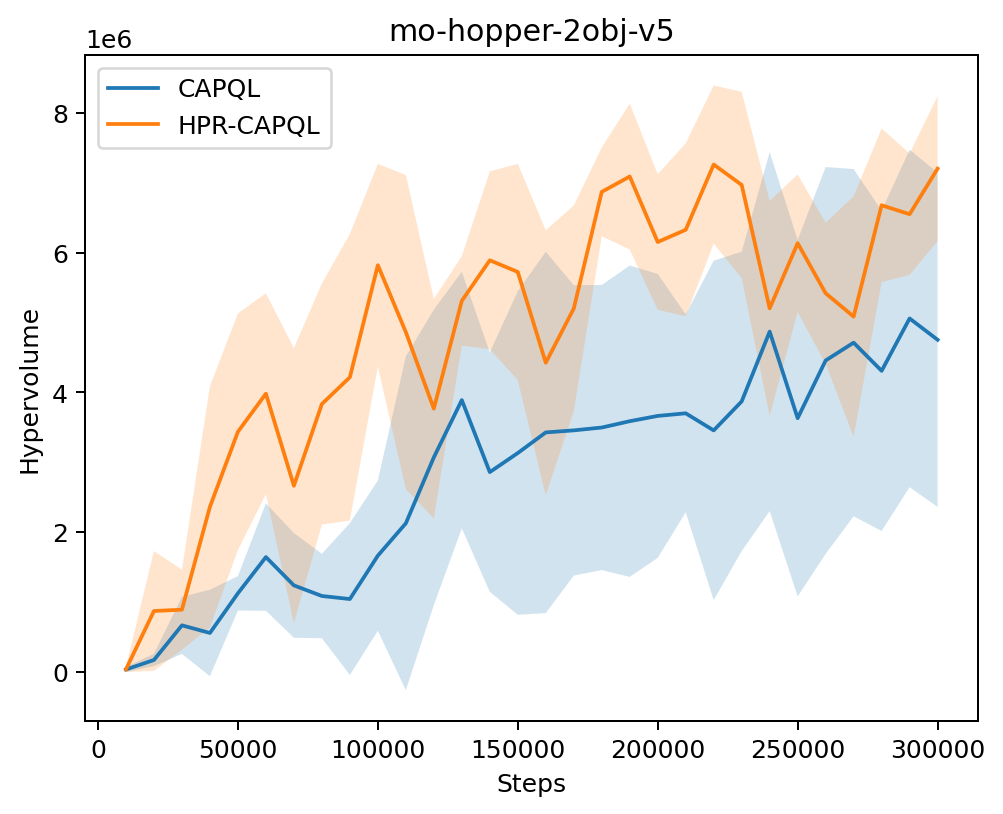}\hfill
  \includegraphics[width=0.32\textwidth]{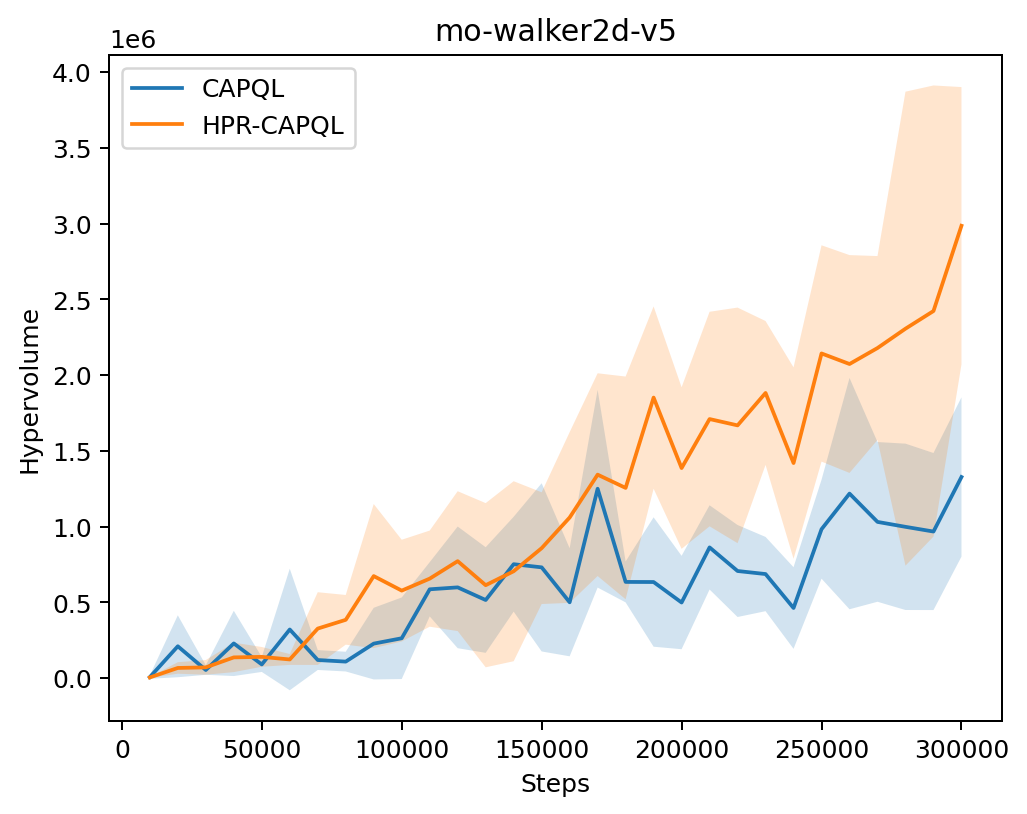}\hfill
  \includegraphics[width=0.32\textwidth]{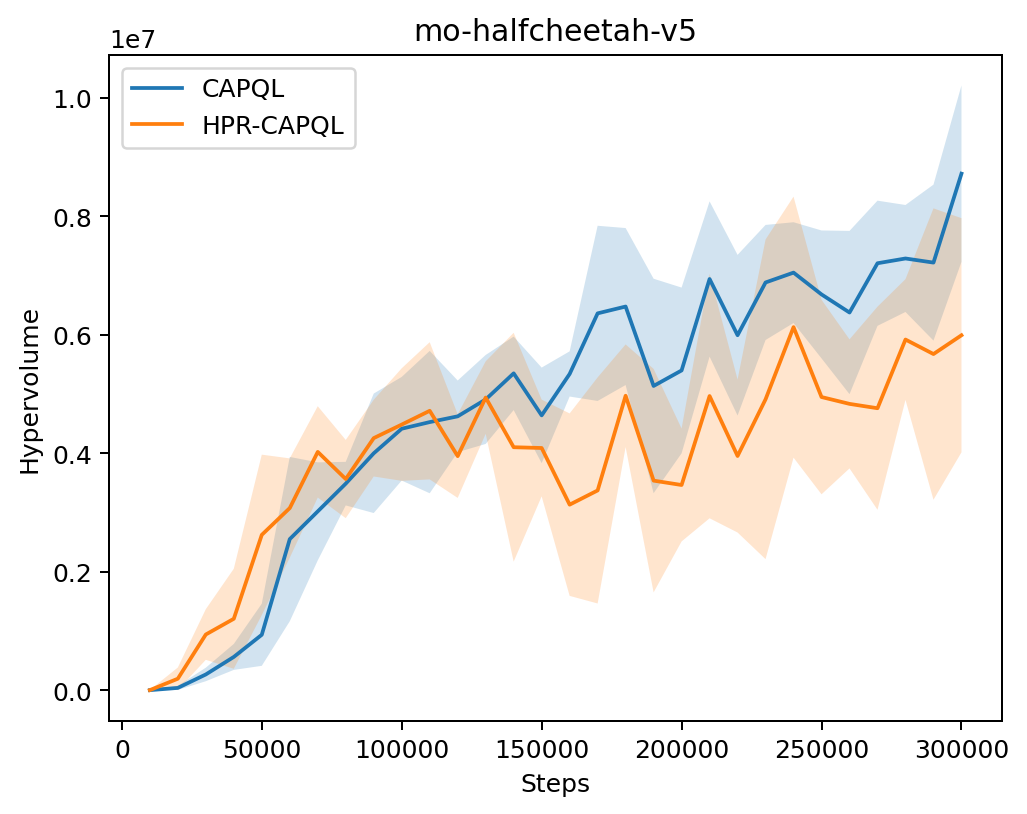}\\[0.6em]
  \includegraphics[width=0.32\textwidth]{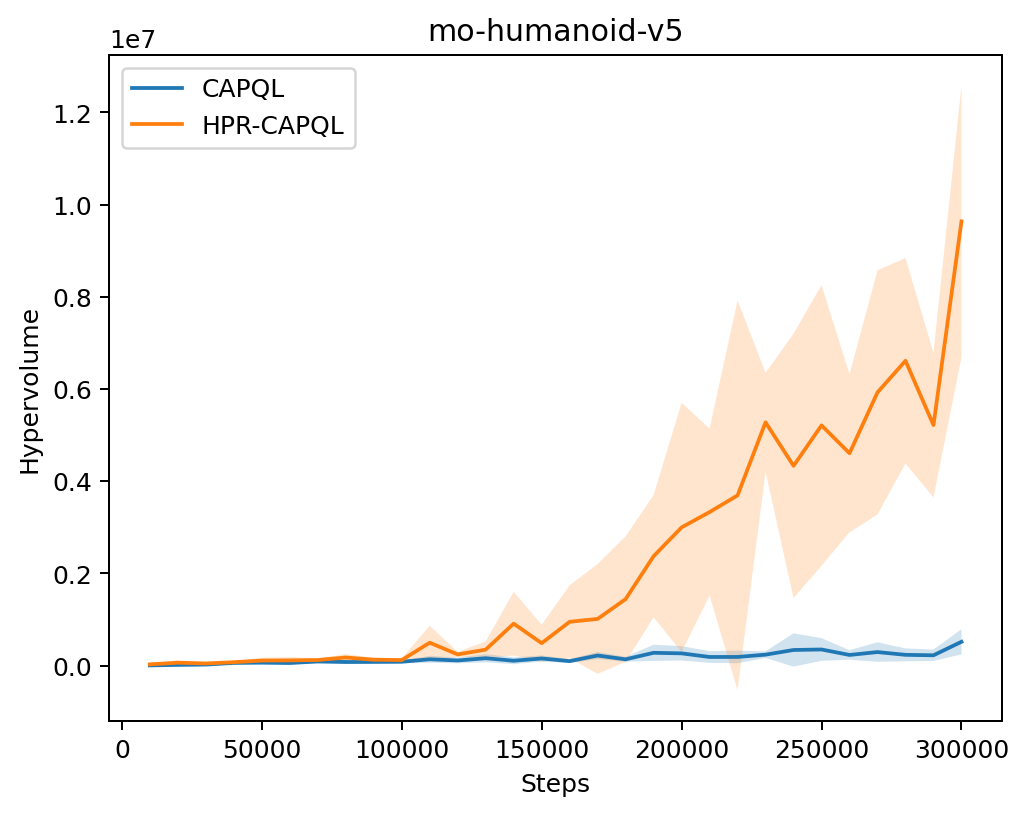}\hfill
  \includegraphics[width=0.32\textwidth]{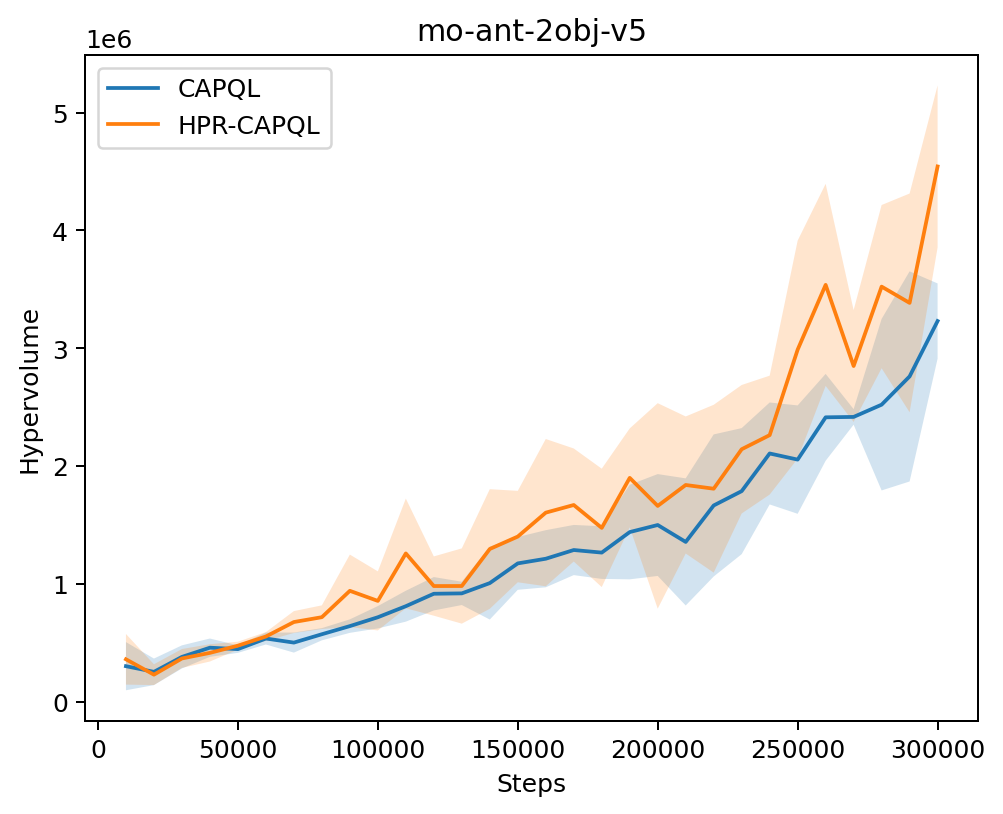}\hfill
  \includegraphics[width=0.32\textwidth]{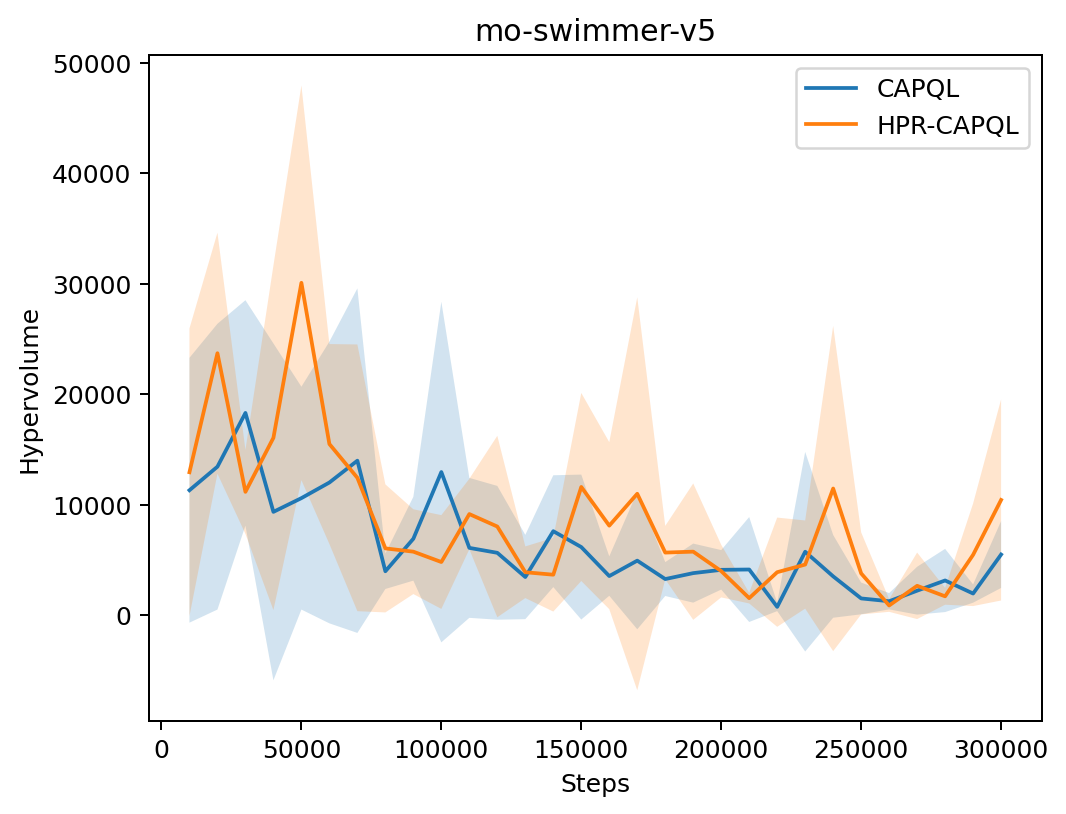}
  \caption{Hypervolume (HV) learning curves up to $300{,}000$ steps (mean $\pm$ std over five seeds).}
  \label{fig:hv-curves}
\end{figure*}
\begin{figure*}[hbt!]
  \centering
  \includegraphics[width=0.32\textwidth]{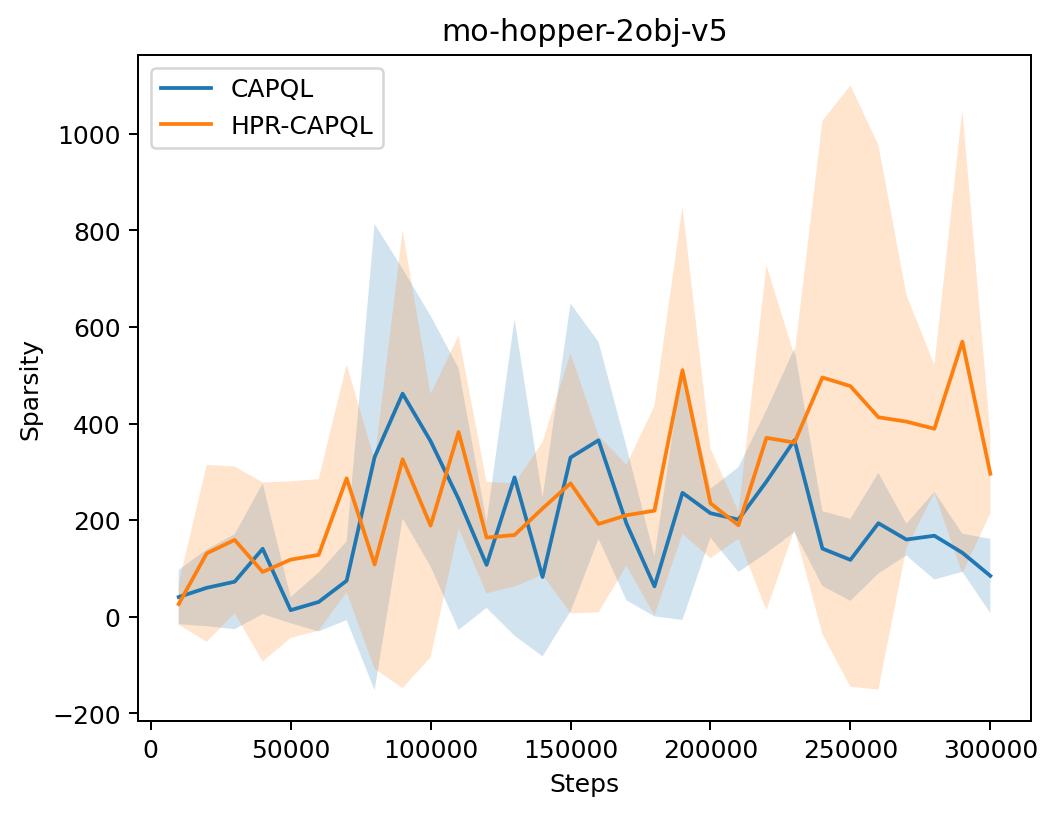}\hfill
  \includegraphics[width=0.32\textwidth]{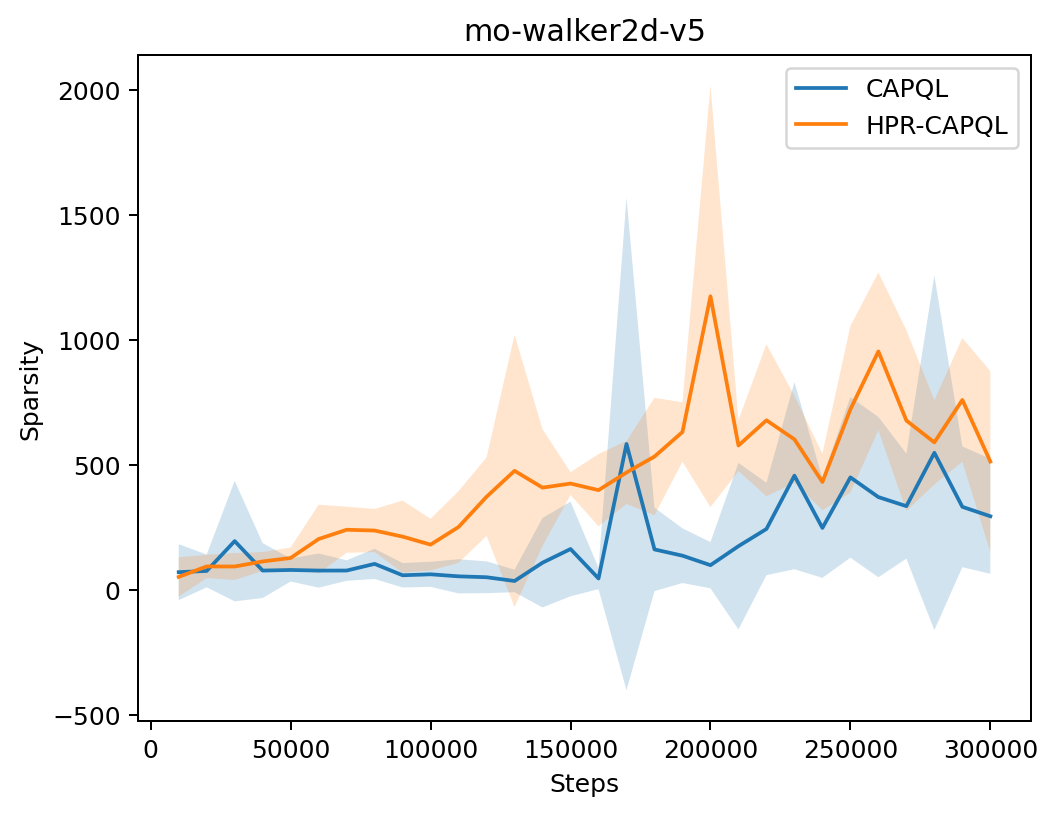}\hfill
  \includegraphics[width=0.32\textwidth]{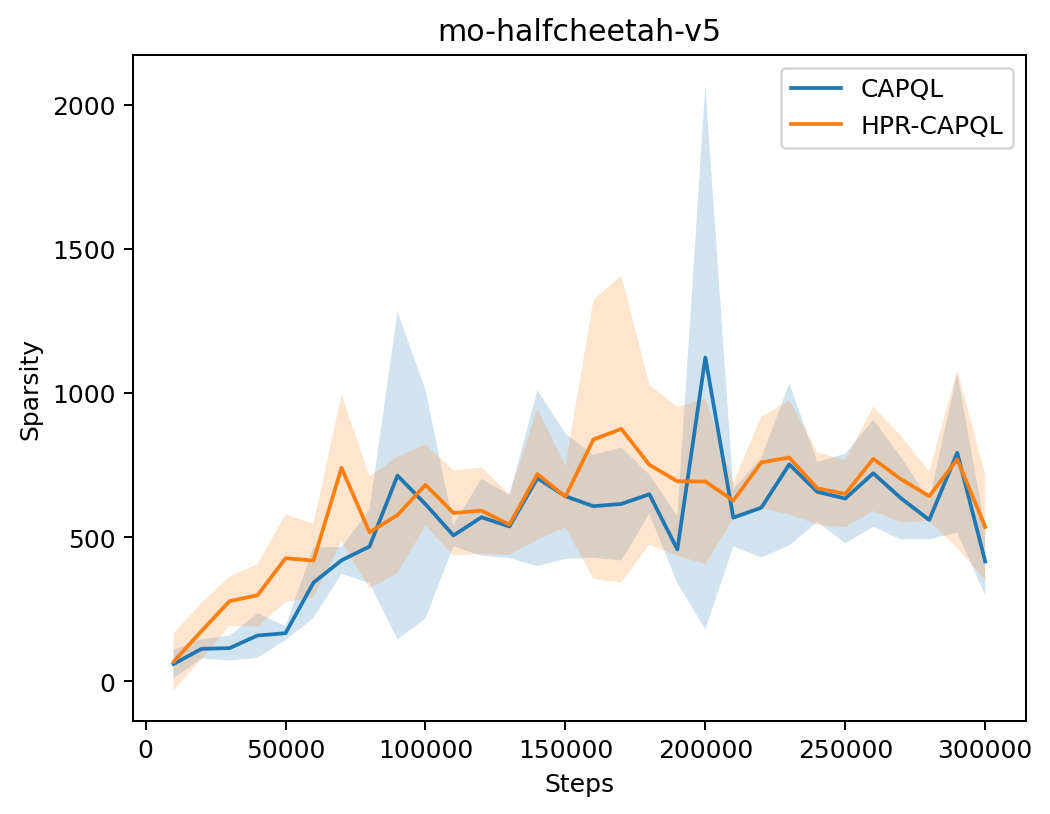}\\[0.6em]
  \includegraphics[width=0.32\textwidth]{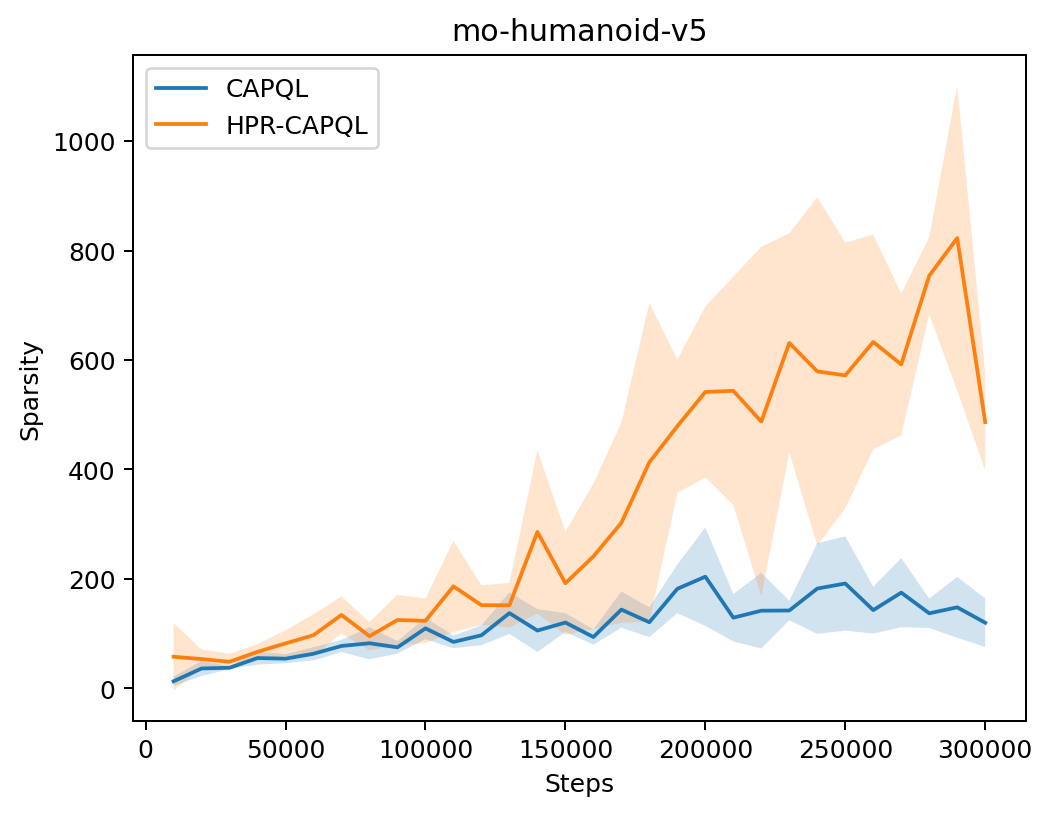}\hfill
  \includegraphics[width=0.32\textwidth]{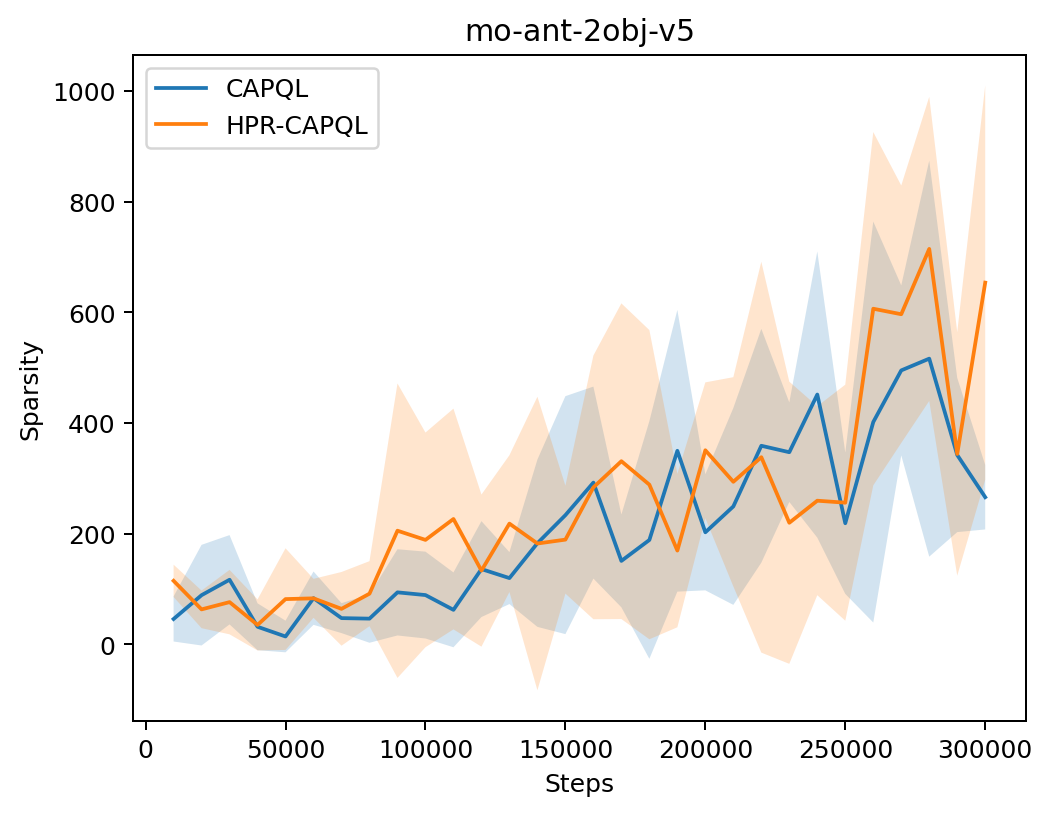}\hfill
  \includegraphics[width=0.32\textwidth]{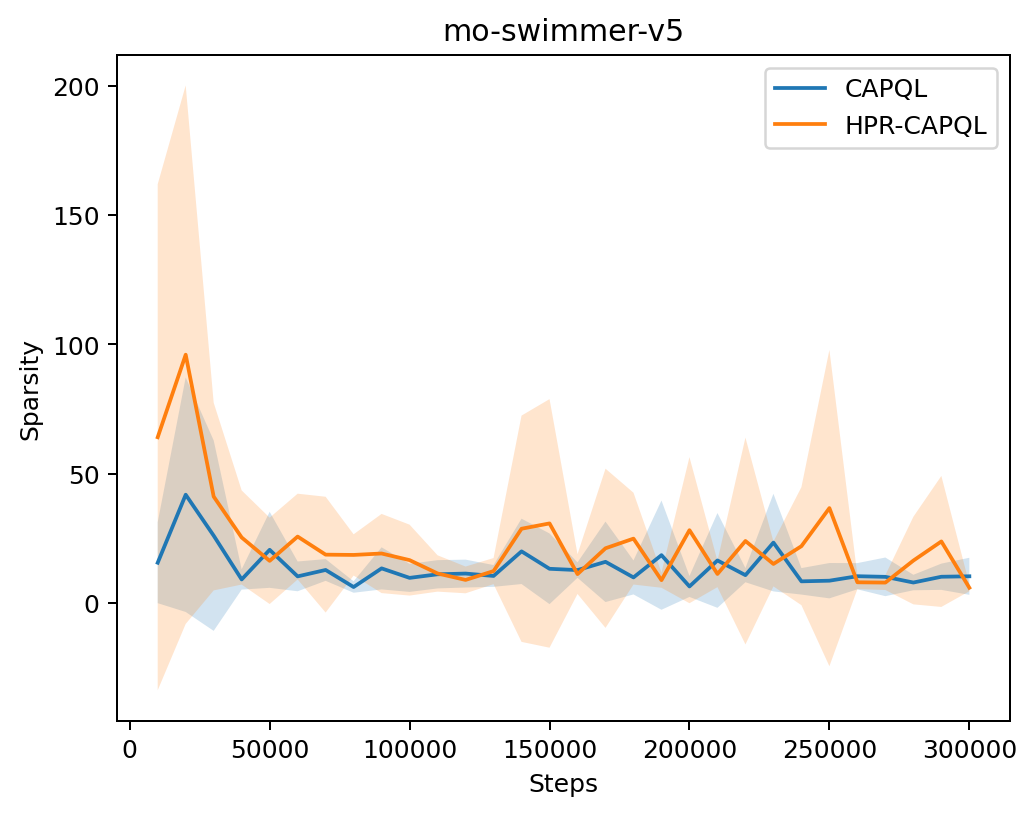}
  \caption{Sparsity (lower is better) up to $300{,}000$ steps (mean $\pm$ std over five seeds).}
  \label{fig:Sparsity-curves}
\end{figure*}

\subsection{Pareto-Front Comparisons}
Figure~\ref{fig:pareto} shows the final non-dominated Pareto fronts (union over seeds at $300{,}000$ steps), revealing where each method extends coverage and how these geometric differences drive the EUM and HV gaps. On \texttt{mo-humanoid-v5}, \texttt{mo-walker2d-v5}, \texttt{mo-hopper-2obj-v5}, and \texttt{mo-ant-2obj-v5}, HPR-CAPQL pushes outward along large portions of the trade-off surface, occupying regions that CAPQL rarely reaches. This outward shift explains both the hypervolume gains and the higher average scalarized utility across preferences. In \texttt{mo-halfcheetah-v5}, CAPQL extends further toward an extreme of one objective, which is sufficient to dominate HV despite similar EUM. In \texttt{mo-swimmer-v5}, the fronts largely overlap, aligning with the inconclusive learning curves.

\begin{figure*}[h]
  \centering
  \includegraphics[width=0.32\textwidth]{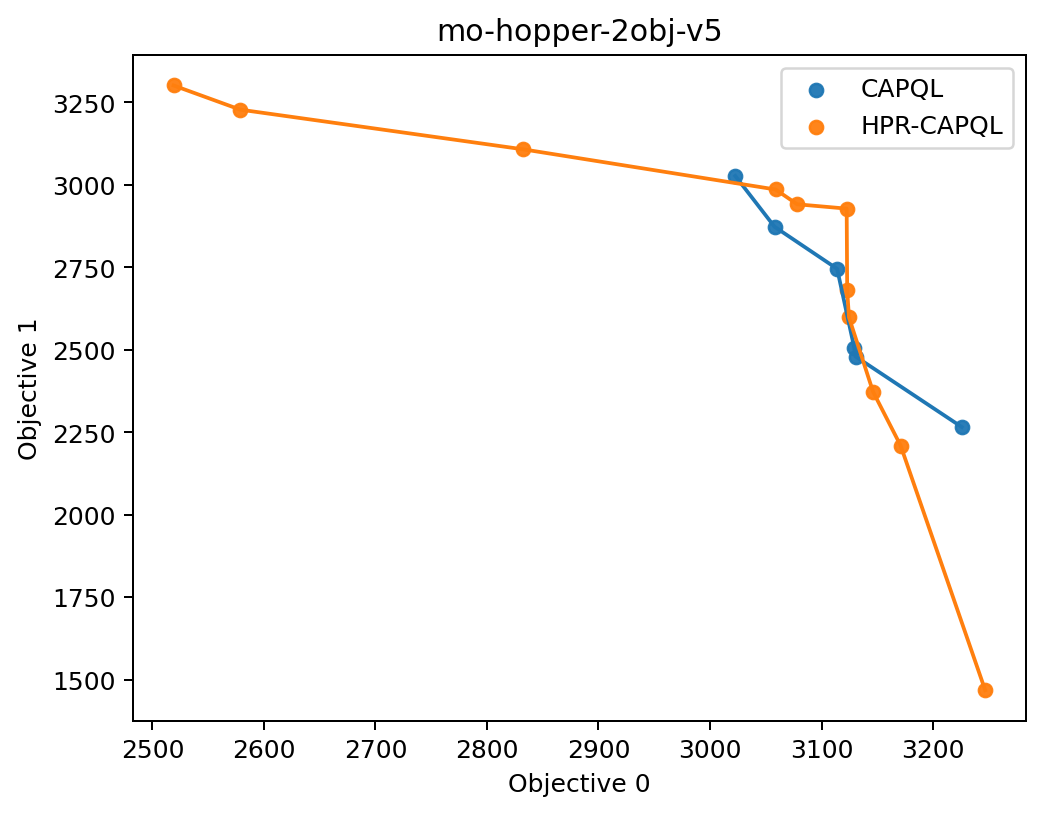}\hfill
  \includegraphics[width=0.32\textwidth]{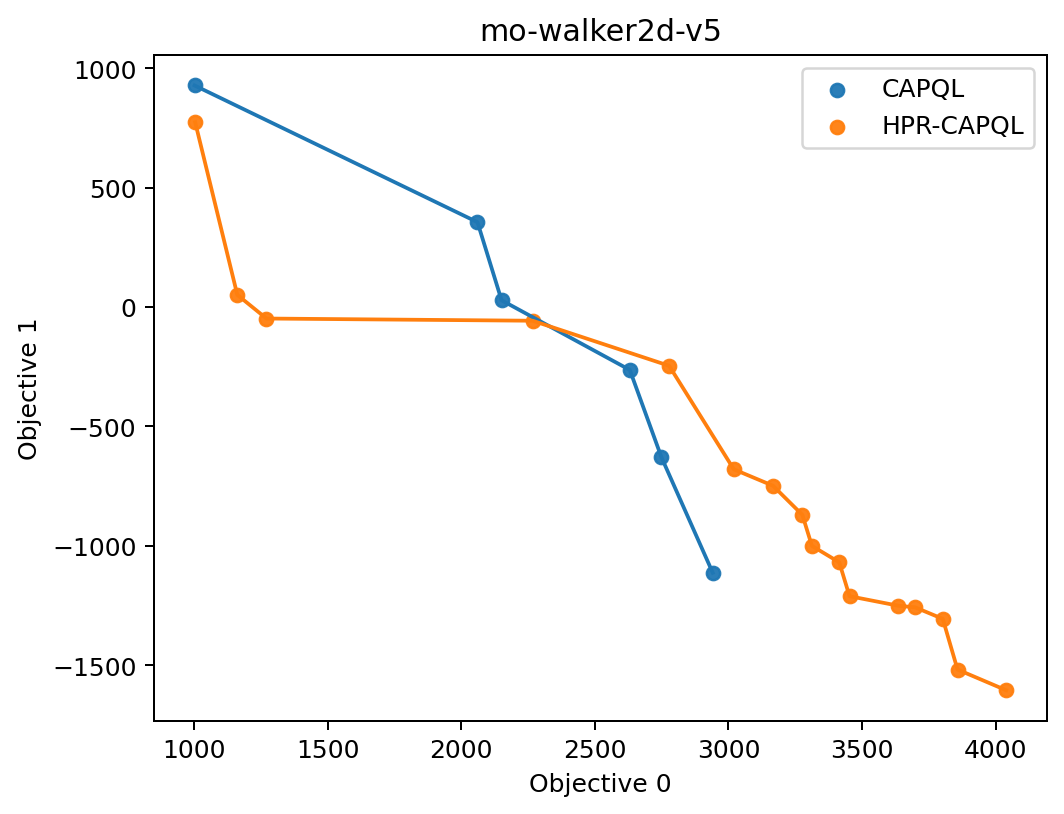}\hfill
  \includegraphics[width=0.32\textwidth]{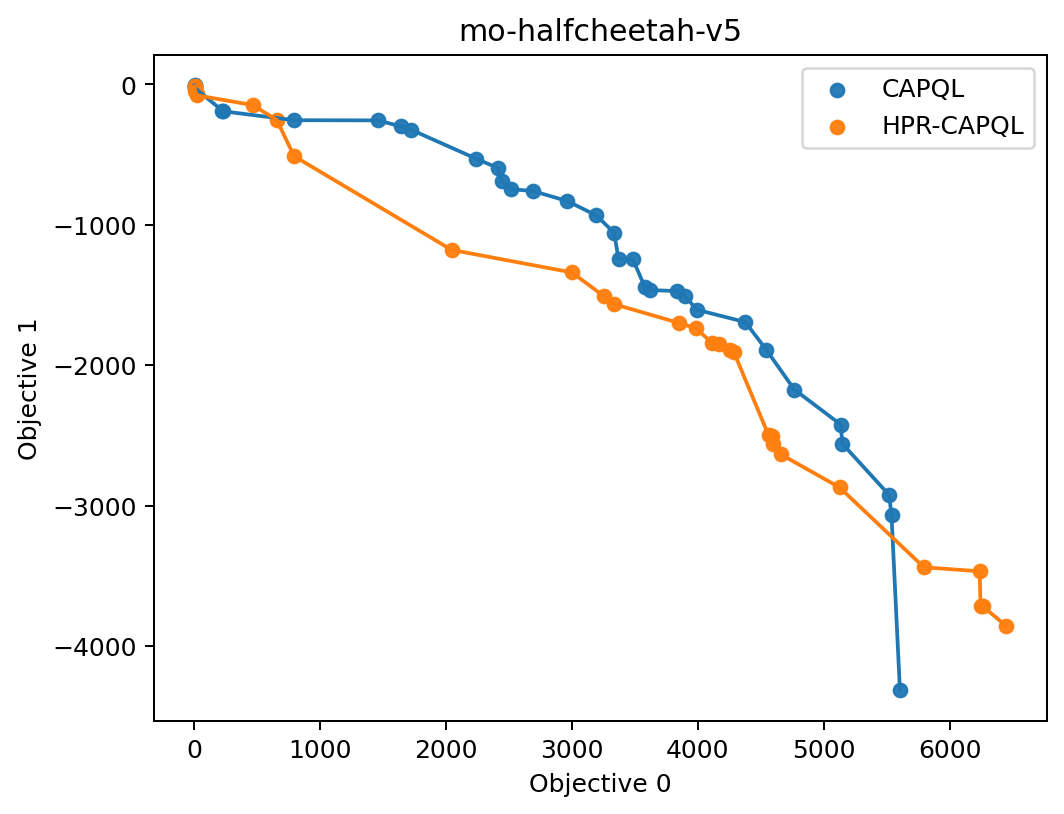}\\[0.6em]
  \includegraphics[width=0.32\textwidth]{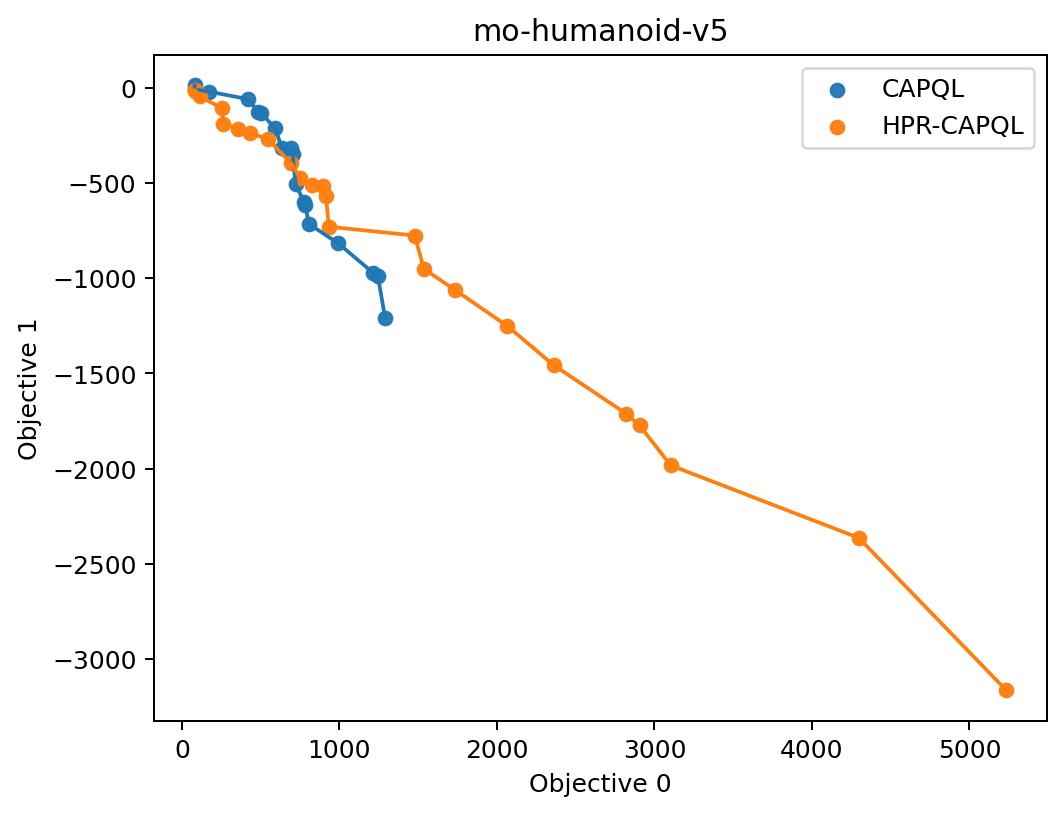}\hfill
  \includegraphics[width=0.32\textwidth]{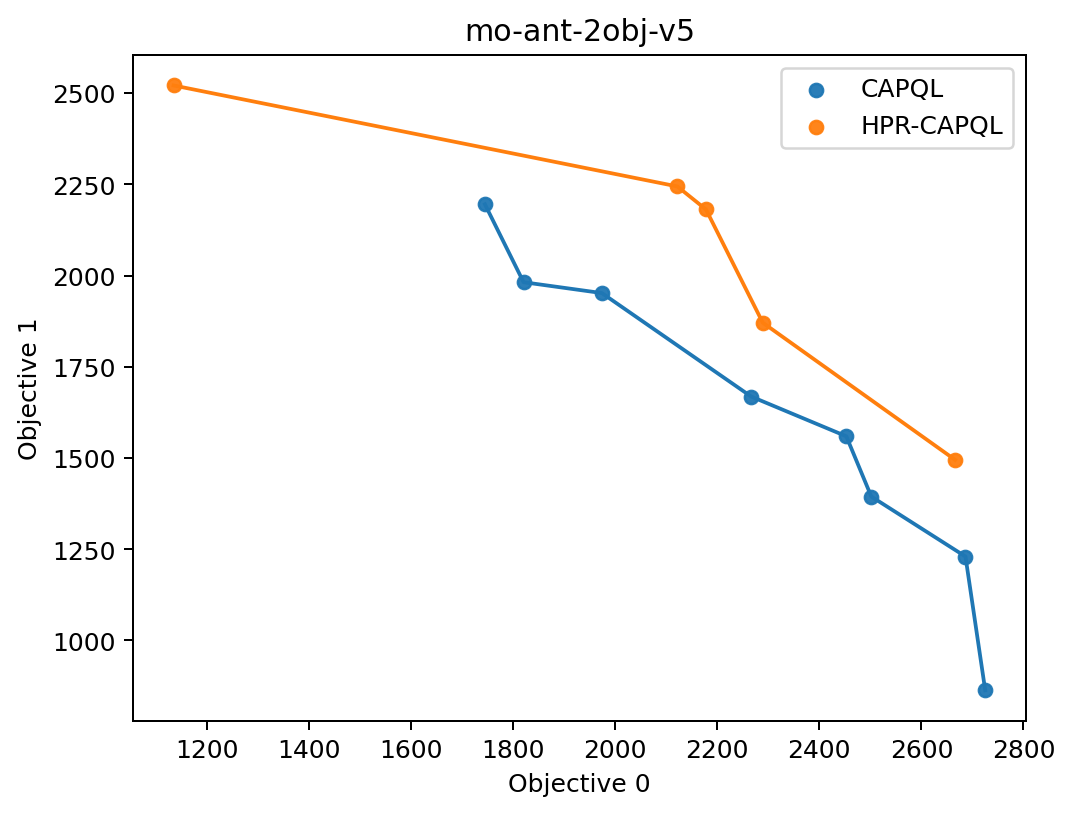}\hfill
  \includegraphics[width=0.32\textwidth]{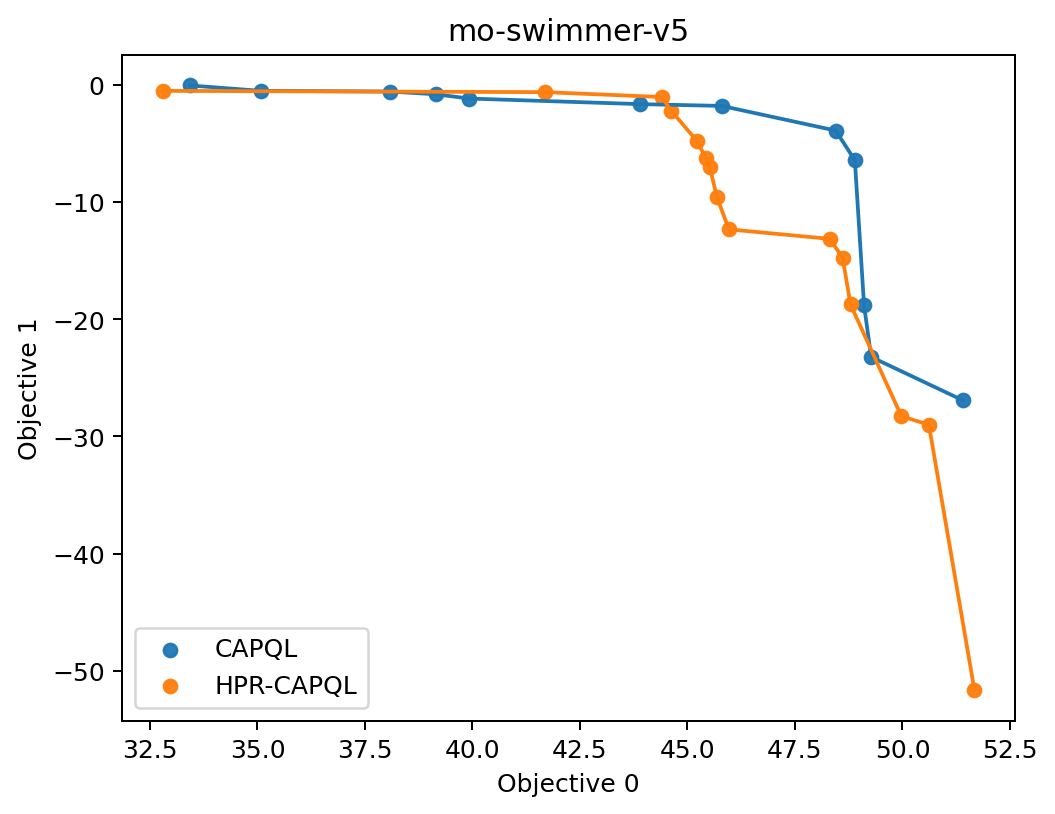}
  \caption{Final Pareto fronts (non-dominated only) after $300{,}000$ steps across all environments. Each panel shows the union of non-dominated solutions across five seeds; legend: CAPQL vs.\ HPR-CAPQL}

  \label{fig:pareto}
\end{figure*}

\subsection{Coverage–Density Trade-off}
Where HPR-CAPQL expands the frontier, sparsity tends to increase: the archive becomes more widely spaced while reaching new territory. This is a natural consequence of a fixed training budget; growing the frontier and densifying it are different objectives. A short densification phase near the end of training, reducing the relabel ratio or temporarily disabling HPR, tightens spacing without relinquishing the newly discovered regions. When dense coverage is paramount, sampling hindsight preferences closer to the uniform simplex or adding small jitter around current preferences further improves spacing.

\subsection{Environment-Specific Remarks}
On \texttt{mo-humanoid-v5}, HPR-CAPQL provides the clearest improvement: the frontier enlarges substantially and both EUM and HV rise throughout the latter half of training. \texttt{mo-walker2d-v5} exhibits steady gains with moderate variance; the final fronts extend outward with only modest loss of density. \texttt{mo-hopper-2obj-v5} shows strong early advantages that persist, with the HPR front lying above the baseline over most of the trade-off range. \texttt{mo-ant-2obj-v5} mirrors this behavior but illustrates the coverage–density trade-off most clearly. \texttt{mo-halfcheetah-v5} favors CAPQL on HV; here, aggressive preference relabeling appears mismatched with the task’s reward geometry. \texttt{mo-swimmer-v5} is simple enough that both methods rapidly converge to similar fronts.

\section{Conclusion}
We presented Hindsight Preference Replay (HPR), a simple replay augmentation that retroactively relabels stored transitions with alternative preference vectors and slots into CAPQL without architectural or loss changes. Under a fixed $300{,}000$-step budget across six MO-Gymnasium tasks, HPR-CAPQL consistently broadens the learned Pareto fronts and often increases preference-averaged utility. In aggregate, hypervolume (HV) is higher in five of six environments and Expected Utility (EUM) in four of six, while CAPQL retains an advantage on \texttt{mo-halfcheetah-v5}.

A few highlights illustrate the magnitude and reliability of the effects. On \texttt{mo-humanoid-v5}, HPR-CAPQL delivers large gains (EUM $1613\!\pm\!464$ vs.\ $323\!\pm\!125$; HV $9.63$M vs.\ $0.52$M) with strong statistical support ($p=0.0009$). \texttt{mo-walker2d-v5} shows a similar pattern (HV $2.99$M vs.\ $1.33$M; $p=0.0079$; EUM $1321\!\pm\!208$ vs.\ $1056\!\pm\!273$), and \texttt{mo-ant-2obj-v5} also favors HPR-CAPQL on both metrics (HV $4.54$M vs.\ $3.23$M; $p=0.0060$; EUM $2061\!\pm\!157$ vs.\ $1823\!\pm\!191$). Two domains temper the story: \texttt{mo-hopper-2obj-v5} trends toward higher HV for HPR-CAPQL but is not significant at $5\%$ ($p=0.0749$), and \texttt{mo-halfcheetah-v5} significantly favors CAPQL on HV (HV $8.72$M vs.\ $5.99$M; $p=0.035$) with similar EUM across methods. \texttt{mo-swimmer-v5} is effectively a draw (small, noisy HV differences; $p=0.2926$).

A consistent pattern is a coverage–density trade-off. When HPR broadens the Pareto frontier, the nondominated archive often becomes more widely spaced, especially where HV gains are largest. This is expected under a fixed budget, since expanding the frontier and densifying it are different goals. A short densification phase near the end, reducing the relabel ratio or briefly disabling HPR, tightens spacing without losing newly reached regions. For tasks such as \texttt{mo-halfcheetah-v5}, conservative relabeling and return-alignment filters further curb variance and harmful updates.

These results support HPR as a strong default for preference-conditioned MORL. It reuses off-policy data across preferences and yields broader Pareto coverage and higher preference-averaged performance with minimal engineering. Future work includes many-objective settings, adaptive preference distributions, learned acceptance criteria, and larger seed budgets.

\section*{Acknowledgements}
This work is supported by the University of Galway College of Science and Engineering Postgraduate Scholarship.

%
%
%
\bibliographystyle{splncs04}
\bibliography{bibliography}

@article{andrychowicz2017hindsight,
  title={Hindsight experience replay},
  author={Andrychowicz, Marcin and Wolski, Filip and Ray, Alex and Schneider, Jonas and Fong, Rachel and Welinder, Peter and McGrew, Bob and Tobin, Josh and Pieter Abbeel, OpenAI and Zaremba, Wojciech},
  journal={Advances in neural information processing systems},
  volume={30},
  year={2017}
}

@article{van2014multi,
  title={Multi-objective reinforcement learning using sets of pareto dominating policies},
  author={Van Moffaert, Kristof and Now{\'e}, Ann},
  journal={The Journal of Machine Learning Research},
  volume={15},
  number={1},
  pages={3483--3512},
  year={2014},
  publisher={JMLR. org}
}

@inproceedings{lu2023multi,
  title={Multi-objective reinforcement learning: Convexity, stationarity and pareto optimality},
  author={Lu, Haoye and Herman, Daniel and Yu, Yaoliang},
  booktitle={The Eleventh International Conference on Learning Representations},
  year={2023}
}

@inproceedings{haarnoja2018soft,
  title={Soft actor-critic: Off-policy maximum entropy deep reinforcement learning with a stochastic actor},
  author={Haarnoja, Tuomas and Zhou, Aurick and Abbeel, Pieter and Levine, Sergey},
  booktitle={International conference on machine learning},
  pages={1861--1870},
  year={2018},
  organization={Pmlr}
}

@article{roijers2013survey,
  title={A survey of multi-objective sequential decision-making},
  author={Roijers, Diederik M and Vamplew, Peter and Whiteson, Shimon and Dazeley, Richard},
  journal={Journal of Artificial Intelligence Research},
  volume={48},
  pages={67--113},
  year={2013}
}

@inproceedings{zitzler2007hypervolume,
  title={The hypervolume indicator revisited: On the design of Pareto-compliant indicators via weighted integration},
  author={Zitzler, Eckart and Brockhoff, Dimo and Thiele, Lothar},
  booktitle={International Conference on Evolutionary Multi-Criterion Optimization},
  pages={862--876},
  year={2007},
  organization={Springer}
}

@inproceedings{auger2009theory,
  title={Theory of the hypervolume indicator: optimal $\mu$-distributions and the choice of the reference point},
  author={Auger, Anne and Bader, Johannes and Brockhoff, Dimo and Zitzler, Eckart},
  booktitle={Proceedings of the tenth ACM SIGEVO workshop on Foundations of genetic algorithms},
  pages={87--102},
  year={2009}
}

@article{mogymnasium,
  title={A toolkit for reliable benchmarking and research in multi-objective reinforcement learning},
  author={Felten, Florian and Alegre, Lucas N and Nowe, Ann and Bazzan, Ana and Talbi, El Ghazali and Danoy, Gr{\'e}goire and C da Silva, Bruno},
  journal={Advances in Neural Information Processing Systems},
  volume={36},
  pages={23671--23700},
  year={2023}
}

@inproceedings{shianifar2024optimizing,
  title={Optimizing Deep Reinforcement Learning for Adaptive Robotic Arm Control},
  author={Shianifar, Jonaid and Schukat, Michael and Mason, Karl},
  booktitle={International Conference on Practical Applications of Agents and Multi-Agent Systems},
  pages={293--304},
  year={2024},
  organization={Springer}
}

@article{gymnasium,
  title={Gymnasium: A standard interface for reinforcement learning environments},
  author={Towers, Mark and Kwiatkowski, Ariel and Terry, Jordan and Balis, John U and De Cola, Gianluca and Deleu, Tristan and Goul{\~a}o, Manuel and Kallinteris, Andreas and Krimmel, Markus and KG, Arjun and others},
  journal={arXiv preprint arXiv:2407.17032},
  year={2024}
}

@article{hayes2021practical,
  title={A practical guide to multi-objective reinforcement learning and planning},
  author={Hayes, Conor F and R{\u{a}}dulescu, Roxana and Bargiacchi, Eugenio and K{\"a}llstr{\"o}m, Johan and Macfarlane, Matthew and Reymond, Mathieu and Verstraeten, Timothy and Zintgraf, Luisa M and Dazeley, Richard and Heintz, Fredrik and others},
  journal={arXiv preprint arXiv:2103.09568},
  year={2021}
}

@article{shianifar2025adaptive,
  title={Adaptive scalarization in multi-objective reinforcement learning for enhanced robotic arm control},
  author={Shianifar, Jonaid and Schukat, Michael and Mason, Karl},
  journal={Neurocomputing},
  pages={132205},
  year={2025},
  publisher={Elsevier}
}

@article{zhang2023multi,
  title={Multi-objective reinforcement learning--concept, approaches and applications},
  author={Zhang, Linzi and Qi, Zhiquan and Shi, Yong},
  journal={Procedia Computer Science},
  volume={221},
  pages={526--532},
  year={2023},
  publisher={Elsevier}
}

@inproceedings{yangpreference,
  title={Preference Controllable Reinforcement Learning with Advanced Multi-Objective Optimization},
  author={Yang, Yucheng and Zhou, Tianyi and Pechenizkiy, Mykola and Fang, Meng},
  booktitle={Forty-second International Conference on Machine Learning (ICML)},
  year={2024}

}

@article{fan2025preference,
  title={Preference-Driven Multi-Objective Combinatorial Optimization with Conditional Computation},
  author={Fan, Mingfeng and Zhou, Jianan and Zhang, Yifeng and Wu, Yaoxin and Chen, Jinbiao and Sartoretti, Guillaume Adrien},
  journal={arXiv preprint arXiv:2506.08898},
  year={2025}
}

@article{wan2018advances,
  title={Advances in experience replay},
  author={Wan, Tracy and Xu, Neil},
  journal={arXiv preprint arXiv:1805.05536},
  year={2018}
}
\end{document}